\documentclass[runningheads]{llncs}
\usepackage[T1]{fontenc}
\usepackage{graphicx}
\usepackage{amsmath,amsfonts}
\usepackage{algorithmic}
\usepackage{algorithm}
\usepackage{array}
\usepackage[caption=false,font=normalsize,labelfont=sf,textfont=sf]{subfig}
\usepackage{textcomp}
\usepackage{stfloats}
\usepackage{url}
\usepackage{verbatim}
\usepackage{graphicx}
\usepackage{cite}
\usepackage{booktabs}
\usepackage{utfsym}
\usepackage{multirow}
\usepackage{color}
\usepackage{wrapfig}
\usepackage{marvosym}
\begin{document}
\title{ESC-MISR: Enhancing Spatial Correlations for Multi-Image Super-Resolution in Remote Sensing}
%
%
\author{Zhihui Zhang\inst{1}\orcidID{0009-0009-5547-3490} \and Jinhui Pang\inst{1}\textsuperscript{(\Letter)}\orcidID{0009-0009-5543-4512} \and Jianan Li\inst{1}\orcidID{0000-0002-6936-9485} \and Xiaoshuai Hao\inst{2}\orcidID{0009-0007-4209-6695} }
\authorrunning{Z. Zhang et al.}
%
\institute{Beijing Institute of Technology, Beijing, China\\
\email{\{3220231441,pangjinhui,lijianan\}@bit.edu.cn} \and
Beijing Academy of Artificial Intelligence, Beijing, China\\
\email{xshao@baai.ac.cn}}
\maketitle              
\begin{abstract}
Multi-Image Super-Resolution (MISR) is a crucial yet challenging research task in the remote sensing community. 
In this paper, we address the challenging task of Multi-Image Super-Resolution in Remote Sensing (MISR-RS), aiming to generate a High-Resolution (HR) image from multiple Low-Resolution (LR) images obtained by satellites. Recently, the weak temporal correlations among LR images have attracted increasing attention in the MISR-RS task. 
However, existing MISR methods treat the LR images as sequences with strong temporal correlations, overlooking spatial correlations and imposing temporal dependencies. To address this problem, we propose a novel end-to-end framework named Enhancing Spatial Correlations in MISR (ESC-MISR), which fully exploits the spatial-temporal relations of multiple images for HR image reconstruction. Specifically, we first introduce a novel fusion module named \textbf{M}ulti-\textbf{I}mage \textbf{S}patial \textbf{T}ransformer (MIST), which emphasizes parts with clearer global spatial features and enhances the spatial correlations between LR images. 
Besides, we perform a random shuffle strategy for the sequential inputs of LR images to attenuate temporal dependencies and capture weak temporal correlations in the training stage. Compared with the state-of-the-art methods, our ESC-MISR achieves 0.70dB and 0.76dB cPSNR improvements on the two bands of the PROBA-V dataset respectively, demonstrating the superiority of our method.

\keywords{Multi-image super-resolution \and high-resolution remote sensing images\and spatial transformer \and spatial-temporal correlations.}
\end{abstract}

\section{Introduction}


Super-resolution algorithms serve the reconstruction of a High-Resolution (HR) image from one or a set of Low-Resolution (LR) images. 
High-Resolution (HR) remote sensing images, derived from advanced remote sensing techniques, have plenty of applications including land cover change analysis, urban planning, and meteorological forecasting. 
Nevertheless, the acquisition of HR images often necessitates costly satellites or alternative data sources because of hardware constraints. To address this challenge, super-resolution technologies continue to develop and have received widespread attention in the realm of remote sensing.

Regarding Super-Resolution in Remote Sensing (SR-RS), research predominantly focuses on Single-Image Super-Resolution in Remote Sensing (SISR-RS) and Multi-Image Super-Resolution in Remote Sensing (MISR-RS). SISR concentrates on reconstructing an HR image from a single LR image. However, remote sensing images exhibit a weaker correlation than general images and necessitate leveraging effective information from multiple scenes. Consequently, MISR has emerged as a vital technology in remote sensing for generating HR images from multiple LR images.




Through extensive research on MISR-RS~\cite{molini2019deepsum,dorr2020satellite,yu2018wide,deudon2019highres}, we note that existing models for MISR-RS are highly sensitive to temporal sequences and neglect spatial correlations among LR images. These methods consider LR images as sequences strongly correlated over time.  
However, factors like long intervals, cloud occlusions, and lighting variations result in weak temporal correlations among LR images in each scene, and LR images contain complementary spatial information. It is essential to weigh up the spatial-temporal characteristics in the process of generating HR images by employing multiple LR images. 
Although RAMS~\cite{salvetti2020multi} and PIU~\cite{valsesia2021permutation} attempt to overcome temporal dependencies, they do not fully eliminate the temporal impact of frame orders. 
TR-MISR~\cite{an2022tr} focuses on all image patches and adapts to multi-image scenarios with weak temporal correlations. However, it does not emphasize the spatial correlations between multiple images.


To address these limitations, we propose a novel end-to-end framework Enhancing Spatial Correlations in MISR (ESC-MISR), tapping complementary spatial features and mitigating temporal dependencies. 
Concretely, we design a new fusion module \textbf{M}ulti-\textbf{I}mage \textbf{S}patial \textbf{T}ransformer (MIST), to enhance spatial correlations between multiple LR images. 
Our MIST consists of several \textbf{M}ulti-\textbf{I}mage \textbf{S}patial \textbf{A}ttention \textbf{B}locks (MISAB), aiming to enhance the spatial correlations between multiple images. 
The swin transformer, as a typical self-attention mechanism, performs self-attention in each window and obtains spatial information of the entire image by moving the window. However, this cannot effectively perceive the global information of each image. Instead of using the window-based approach, we introduce message tokens~\cite{ni2022expanding} in our MISAB to obtain global spatial information directly. 
The MISAB first calculates the cross-attention of all patches containing multiple pixels within each image. During this period, there is no information exchange between multiple images to weaken temporal correlations. Moreover, after independently obtaining global cross-attention information of each image, we fuse the spatial features of multiple images to mine their complementary spatial information. 

For encoding networks, the vision transformer~\cite{yuan2021incorporating,liang2022vrt,liu2024customized} has already demonstrated strong feature encoding capabilities in super-resolution. Nonetheless, most researchers still prioritize Convolutional Neural Networks (CNNs) for feature encoding in MISR-RS. 
The CNNs-Meet-Transformers (CMT) has proven its stronger encoding ability than CNNs and pure transformer networks such as ViT~\cite{dosovitskiy2020image} and swin transformer~\cite{liu2021swin}. 
Thus, we explore the CMT~\cite{guo2022cmt} to encode LR images better in our ESC-MISR.
Additionally, we employ the Fast Fourier Convolution (FFC)~\cite{chi2020fast} in the decoder to increase the receptive field of the images.

Simultaneously, we employ a random shuffle strategy for LR images of each scene in the training stage, which makes the model insensitive to weak temporal information and generates stable SR images for different orders. 
The experimental results clearly demonstrate the effectiveness of our proposed method in improving the MISR-RS performance. Specifically, ESC-MISR is 0.70dB-1.87dB higher than the state-of-the-art methods on the NIR band, and there is a significant improvement of 0.76dB-1.63dB on the RED band. 


The contributions of this paper are mainly four-fold: 
\begin{itemize}

\item We propose a new framework \textbf{E}nhancing \textbf{S}patial \textbf{C}orrelations in MISR (ESC-MISR) to excavate spatial-temporal relations of multiple Low-Resolution (LR) images, strengthening spatial features while weakening temporal correlations among LR images. 


\item To enhance spatial correlations between LR images, we design a novel fusion module, \textbf{M}ulti-\textbf{I}mage \textbf{S}patial
\textbf{T}ransformer (MIST), which fully utilizes global spatial features for High-Resolution (HR) image reconstruction. 

\item To attenuate temporal dependencies, a random shuffle strategy is implemented for the sequence of multiple LR images to assist our ESC-MISR in fitting weak temporal correlations in the training stage.

\item Compared with the state-of-the-art MISR-RS methods, ESC-MISR obtains 0.65dB and 0.71dB cPSNR improvements on the two bands of the PROBA-V dataset respectively, demonstrating the superiority of our method.
\end{itemize}

\section{Related Work}
\label{sec:related}


\subsection{Single-Image Super-Resolution in Remote Sensing}
SISR techniques aim to construct a High-Resolution image from a single Low-Resolution input. 
Since Dong et al.~\cite{dong2014learning} propose the first CNN-based Super-Resolution (SR) framework, Super-Resolution CNN (SRCNN), a series of SISR methods of deep neural networks have been proposed~\cite{Chen2022ActivatingMP,ju2023resolution}. In the field of remote sensing, ~\cite {dong2019transferred, gu2019deep} integrate residual connections in the model to improve the quality of Low-Resolution bands in multi-spectral remote sensing images. After the popularity of GAN, several variations of GAN such as ESRGAN~\cite{wang2018esrgan} can improve the super-resolution performance of remote sensing images by utilizing perceptual learning and multi-level adaptive weighted fusion. For recent methods, EDiffSR~\cite{Xiao2023EDiffSRAE} and ~\cite{wu2023lightweight} attempt different approaches to further improve the performance of SISR-RS. 


Recently, since the successful application of transformers in the vision field ~\cite{liang2022vrt,ju2023resolution}, more and more SISR works have utilized various transformer variants~\cite{liu2021swin,guo2022cmt} to enhance model feature expression capabilities. In this work, we use the CMT~\cite{guo2022cmt} as our encoder.
However, SISR can only obtain information from a single image, which is insufficient for generating HR remote sensing images, while MISR can better reconstruct HR remote sensing images by capturing effective features from multiple views.



\subsection{Multi-Image Super-Resolution in Remote Sensing}

After the Proba-V~\cite{martens2019super} dataset is released by the European Space Agency, there has been increasing research interest in MISR-RS. DeepSUM~\cite{molini2019deepsum} is the earliest method to use 3D convolution to fuse information from multiple images. 
RAMS~\cite{salvetti2020multi} introduces a new 3D convolutional feature attention mechanism, which focuses the network on HR information and largely overcomes the main local limitations of convolutional operations. In addition, they average the SR features from multiple times of input scrambling, but the model is still sensitive to the temporal information from LR images. 
TR-MISR~\cite{an2022tr} proposes a feature rearrangement module that can not only focus on all patches of the image but also adapt to multiple images with weak temporal correlations. However, this method does not highlight the spatial correlations between multiple images. PIU~\cite{valsesia2021permutation} proposes a method for estimating the uncertainty of SR image arbitrariness based on the variability of input LR images. It improves the quality of SR images from the perspective of generated results. Unfortunately, the model overlooks the influence of frame orders and simply averages the inputs of multiple images into a single image. In contrast, we focus on the sequence of the input and utilize the shuffle strategy to make the model suitable for weak temporal correlations. 
Besides, we note a bias toward extracting global spatial features in the fusion module to enhance spatial correlations.

\section{Methodology}
\label{sec:method}

\subsection{The Overall Structure}
\begin{figure*}[t]
    \centering
    \includegraphics[width=0.9\linewidth]{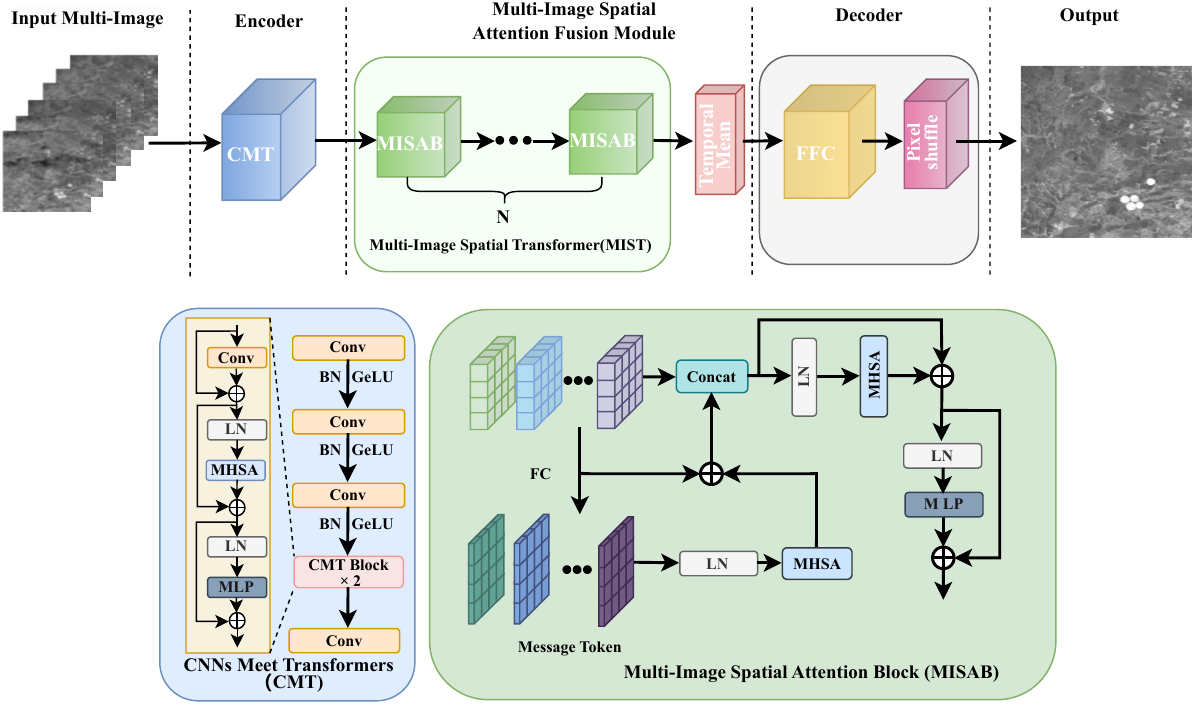}
    \captionsetup{font=small}
    \caption{\textbf{The overall framework of our ESC-MISR.} It consists of three parts: encoder CNNs-Meet-Transformers (CMT), fusion module \textbf{M}ulti-\textbf{I}mage \textbf{S}patial
\textbf{T}ransformer (MIST), and decoder Fast Fourier Convolution (FFC). Additionally, we employ the shuffle strategy to attenuate temporal dependencies in the training stage.}
     \vspace{-0.5cm}
        \label{fig:MISR}
\end{figure*}

Our ESC-MISR employs an encoder-fusion-decoder structure.  For the inputs of LR images $\{\bold\mathcal{I}^i_{LR} \}_{i=1}^K $ $\in$ $R^ {K\times H\times W\times C_{in}}$, we utilize the CNNs-Meet-Transformers (CMT)~\cite{guo2022cmt} to encode feature, where $H, W$ and $C_{in}$ are the height, width, and number of input channels, respectively, and $K$ indicates the number of LR images we specify for each scene, as illustrated in Fig.~\ref{fig:MISR}. The feature maps of images after CMT encoding $\bold\mathcal F_E^i$ $\in$ $R^{H\times W\times C}$ will be fed into our fusion module, \textbf{M}ulti-\textbf{I}mage \textbf{S}patial \textbf{T}ransformer (MIST). MIST highlights spatial information and aims to obtain fusion features $\bold\mathcal F_{SE}$ $\in$ $R^{H\times W\times C}$, where $C$ indicates the channel number of the intermediate feature. 
The fused features are processed by the decoder Fast Fourier Convolution (FFC)~\cite{chi2020fast} to receive decoded feature $\bold\mathcal F_D$ $\in$ $R^{H\times W\times C_{out}}$, which are then up-sampled by the pixel-shuffle~\cite{shi2016real} module to obtain a Super-Resolution (SR) image, 
where $C_{out}$ is the number of output channels, $H^{'}$ and $W^{'}$ are the height and width. Next, we will provide a detailed introduction to the encoder, decoder, and fusion module of ESC-MISR. The shuffle strategy on ESC-MISR will be presented in the training stage.
    

\subsection{Encoder and Decoder of ESC-MISR}
\noindent
\textbf{The CNNs-Meet-Transformers (CMT) Encoder.} Since each set in the PROBA-V dataset has different numbers of LR images, we specify the input image number as $K$. If the input $\{I^i_{LR} \}_{i=1}^N$ $\in$ $R^{B\times N\times H\times W\times C}$ 
are less than $K$ frames, we operate $Max(LR_{clr}^1, LR_{clr}^2,..., LR_{clr}^N)$ to select the clearest image as padding image $I_{pd}$, whose quantity is $K-N$, where $LR_{clr}$ is the clarity of LR images mentioned in Sec.~\ref{sec:dataset}, $B$ means batch size, and $N$ indicates the number of LR images contained in each scene. 

The CMT combines CNN and transformer and has four stages for encoding images. We can achieve the desired effect at a lower cost through only the first stage of the feature processing. 
Therefore, we employ the first-stage CMT to extract deep features $\{\bold\mathcal F_E^i\}_{i=1}^K$ $\in$ $R^{B\times K\times H\times W\times C}$ of the input $\{\bold\mathcal I^i_{LR} \}_{i=1}^K$, which first pass through three convolution layers with the BatchNorm (BN) layer and GeLU~\cite{hendrycks2016gaussian} activation layer. As demonstrated in Fig.~\ref{fig:MISR}, the entire encoding process of CMT for input feature $\bold\mathcal{X}$ as: 
\begin{equation}
    {{\rm CMT}}(\bold\mathcal {X})={\rm Conv}({\rm CMTB}(\bold\mathcal X_C)),
\end{equation}
where $\bold\mathcal X_C={\rm BN}({\rm GeLU}({\rm Conv}(\bold\mathcal X)))$. ${\rm Conv}(\cdot) $ denotes a $3\times 3$ convolution layer. Meanwhile, a CMT block (CMTB) consists of ${\rm Conv}(\cdot) $, Multi-Head Self-Attention (MHSA) with LayerNorm (LN) layer and Multi-Layer Perceptron (MLP) layer. For shallow feature $\bold\mathcal{X_C}$, CMTB conducts deep feature extraction as:
\begin{equation}
    \bold\mathcal {X_C}^{'}=\bold\mathcal {X_C} + {\rm Conv}(\bold\mathcal {X_C}),
\end{equation}
\begin{equation}
   \bold\mathcal {X_A}=\bold\mathcal {X_C}^{'} + {\rm MHSA}({\rm LN}(\bold\mathcal {X_C}^{'})).
\end{equation}

For the MHSA, the key, query, and value matrices are acquired by multiplying the matrix of intermediate feature $\bold\mathcal X_C^{'}$ after processing through the LN layer by the weight matrices $\bold\mathcal W_q$, $\bold\mathcal W_k$, and $\bold\mathcal W_v$ as $\bold\mathcal Q,\bold\mathcal K$, and $\bold\mathcal V$, and it is calculated as:
\begin{equation}\label{MHSA}
    {\rm {Attention}}(\bold\mathcal Q, \bold\mathcal K, \bold\mathcal V)={\rm Softmax}(\frac{\bold\mathcal {QK}^T}{\sqrt{d_k}}+\bold\mathcal B)\bold\mathcal V,
\end{equation}
where $\bold\mathcal B$ signifies the relative position bais, which is randomly initialized and learnable, and $d_k$ represents the dimension of the key.
Afterwards, the intermediate feature $\bold\mathcal X_A$ is calculated by the MLP with the LN layer as:
\begin{equation}
    \bold\mathcal X_M=\bold\mathcal X_A + {\rm MLP}({\rm LN}(\bold\mathcal X_A)),
\end{equation}
to attain the CMT encoded feature $\bold\mathcal F_E={\rm Conv}(\bold\mathcal X_M)$, the MIST, which is built up by several \textbf{M}ulti-\textbf{I}mage \textbf{S}patial \textbf{A}ttention \textbf{B}lock (MISAB), extracts its spatial enhancement fusion feature $\bold\mathcal F_{SE}$ $\in$ $ R^{B\times H\times W\times C}$. The explicit structure of MISAB will be elaborated in Sec.~\ref{SEAB}.

\noindent
\textbf{Fast Fourier Convolution (FFC) Decoder.}
After obtaining fused feature $\bold\mathcal F_{SE}$$ \in R^{B\times H\times W\times C}$, we utilize Fast Fourier Convolution (FFC)~\cite{chi2020fast} for feature decoding to better capture the receptive field of the whole image, which is formulated as:
\begin{equation}
    \bold\mathcal {F}_D={\rm Conv}({\rm FFC}(\bold\mathcal F_{SE})),
\end{equation}
where $\bold\mathcal {F}_{SE}$ falls into the $[\bold\mathcal{f}_l,\bold\mathcal{f}_g]$ implying local part and global part. 
As depicted in Fig.~\ref{fig:MISR}, the thorough operation process of FFC is computed as:
\begin{equation}
    \bold\mathcal{X}_l={\rm ReLU}({\rm BN}(\bold\mathcal{f}_{\mathit{l\to l}} +\bold\mathcal{f}_{\mathit{g\to l}} ),
\end{equation}
\begin{equation}
    \bold\mathcal{X}_{g}={\rm ReLU}({\rm BN}(\bold\mathcal{f}_{\mathit{l\to g}} +\bold\mathcal{f}_{\mathit{g\to g}} ),
\end{equation}
\begin{equation}
    \bold\mathcal{f}_{g\to g}=\bold\mathcal{f}_{g}+{\rm IFFT}({\rm CONV}({\rm FFT}(\bold\mathcal{f}_{\mathit{g}}))),
\end{equation}
where $\bold\mathcal{f}_{l\to l}$ and $\bold\mathcal{f}_{l\to g}$ are obtained by ${\rm Conv}(\bold\mathcal{f}_\mathit{l})$, while $\bold\mathcal{f}_{g\to l}$ are attained by ${\rm Conv}(\bold\mathcal{f}_\mathit{g})$. ${\rm CONV(\cdot)}$ represents ${\rm Conv}({\rm ReLU}({\rm Conv}(\cdot)))$. Specifically, FFT is 2D Fast Fourier Transform and IFFT is Inverse of FFT. Therefore, we can obtain feature $\bold\mathcal{X}_{FFC}=[\bold\mathcal{X}_l,\bold\mathcal{X}_g]$. Before $\bold\mathcal{X}_{FFC}$ is up-sampled by the pixel-shuffle module~\cite{shi2016real}, ${\rm Conv}(\cdot)$ develops $\bold\mathcal{X_\mathit{FFC}}$ $\in$ $ R^{B\times H\times W\times C}$ into the decoded feature $\bold\mathcal{F}_{D}$. 


\subsection{\textbf{M}ulti-\textbf{I}mage \textbf{S}patial \textbf{A}ttention \textbf{B}lock (MISAB)}\label{SEAB}
{To enhance the spatial correlations among multiple images}, we introduce a novel fusion block MISAB for preference spatial features from the perspective of pixels. Our MISAB calculates the cross-attention of each patch feature within each image and integrates the information. 
The swin transformer performs self-attention in each window and obtains spatial information of the entire image by moving the window. However, this approach cannot effectively perceive the global information of each image. Instead of using the window-based approach, we introduce message tokens in our MISAB to obtain global spatial information directly inspired by ~\cite{ni2022expanding}.

We first calculate the message token for each patch, which includes $n$ pixels. Therefore, the CMT encoded feature of each image $\bold\mathcal F_E^i$ is separated into $\{\bold\mathcal F_P^j\}_{j=1}^{N}$, where $N=\frac{H\times W}{n}$, as displayed in Fig.~\ref{fig:MISR}. After linear transformation of patch tokens $\{\bold\mathcal F_P^j\}_{j=1}^{N}$, message tokens $\{\bold\mathcal m_p^j\}_{j=1}^{N}$ are obtained. Then, these tokens are applied to MHSA with LN as:
\begin{equation}
    \bold\mathcal M_P^j=\bold\mathcal m_p^j + {\rm MHSA}({\rm LN}(\bold\mathcal m_\mathit{p}^\mathit{j})),
\end{equation}
where $j=1,...,N$, and the message token $\bold\mathcal M_I^i$ carried by each image feature $\bold\mathcal F_E^i$ involves $N$ patches message tokens $\{\bold\mathcal M_P^j\}_{j=1}^N$. Afterwards, feature $\bold\mathcal F_M^i$ contains spatial cross-attention information, which concatenates the two features as $[\bold\mathcal F_E^i,\bold\mathcal M_I^i]$. Then, the features $\{\bold\mathcal F^i_M\}_{i=1}^K$ to be fused are represented by $\bold\mathcal{\hat{F}}_M$. Unlike calculating global spatial-temporal information in ~\cite{ni2022expanding}, this step weakens temporal information and enhances attention to spatial features of pixels, instead of involving information interaction between frames.


    

Next, we employ the \textbf{M}ulti-\textbf{I}mage \textbf{S}patial \textbf{A}ttention (MISA) on feature $\bold\mathcal{\hat{F}}_M$ to utilize the multi-image interactive spatial message. This process can be expressed as:
\begin{equation}
    \hat{\bold\mathcal{F}}_{SE}=\bold\mathcal {\hat{F}}_M + {\rm MISA}(\bold\mathcal {\hat{F}}_\mathit{M}),
\end{equation}
where MISA is composed of the LN and MHSA. The MHSA is calculated as Eq.~\ref{MHSA} and the MISA is computed as:
\begin{equation}
    \bold\mathcal Y=\bold\mathcal{\hat{F}}_M + {\rm MHSA}({\rm LN}(\bold\mathcal{\hat{F}}_\mathit{M})),
\end{equation}
\begin{equation}
    \hat{\bold\mathcal{F}}_{SE}=\bold\mathcal Y +  {\rm MLP}({\rm LN}(\bold\mathcal Y)),
\end{equation}
where $\bold\mathcal Y$ indicates an intermediate feature, the spatial enhancement feature $\hat{\bold\mathcal{F}}_{SE}$ incorporates cross information of fused multi-image, which contains $\{\bold\mathcal f_{SE}^i\}_{i=1}^K$. Then, we obtain fused feature ${\bold\mathcal{F}}_{SE}=\frac{1}{K}\sum_{i=1}^K \bold\mathcal f_{SE}^i$. Particularly, OCA~\cite{Chen2022ActivatingMP} also takes a pixel perspective, but it uses pixel tokens to calculate cross-attention within each window feature. Instead of using window-based thought~\cite{liu2021swin}, we innovate message tokens of $\frac{H\times W}{n}$ patches containing $n$ pixels in each image to calculate the cross-attention of these message tokens and allow each image to patch spatial enhancement message. Then, in the multi-image fusion module, MISA is capable of tilting towards spatial information.


\subsection{Multi-Image Shuffle Strategy}
Many MISR-RS models~\cite{yu2018wide,deudon2019highres,molini2019deepsum,dorr2020satellite,an2022tr} are sensitive to the order of multiple LR images. When multi-images are shuffled, the quality of the SR images deteriorates. Multiple LR images are weakly temporally correlated. However, these models mistakenly view multiple LR images as strong temporal correlation sequences. This results in models focusing on temporal information between multiple LR images. To tackle this problem, our idea is that if the model is capable of stably generating high-quality SR images when faced with multi-image inputs in different orders, the robustness of the model can be improved. {Therefore, we introduce the random shuffle strategy to assist ESC-MISR in attenuating the dependency of multi-image input orders and capturing weak temporal correlations in the training stage.}
We employ the random shuffle strategy to arrange multiple images in a stochastic sequence. For the sequence $[I_{LR}^1,I_{LR}^2,...,I_{LR}^K]$, we sort them into $[I_{LR}^i,I_{LR}^j,...,I_{LR}^k]$ after shuffling, where $i, j$, $k$ $\in [1,K]$, and they are not equal to each other. We attempt random shuffle 1 to T times for each multi-image set $\{I^i_{LR} \}_{i=1}^K$ in the experiments. The training time increased by T accordingly. The results will be discussed in detail in Sec.~\ref{ablation}. 
Our ESC-MISR is optimized on the basis of the L2~\cite{dong2016accelerating} loss function. 

\section{Experiments}
\label{sec:experiments}

\subsection{The PROBA-V Kelvin Dataset}
\label{sec:dataset}


The PROBA-V dataset consists of 1450 scenes that cover the RED and NIR spectrum bands from 74 regions on the earth. Of the 1450 scenes, 290 scenes are used for testing. Each scene includes a 384$\times$384 HR image and 9 to 35 128$\times$128 LR images. In addition, there is a clearance.npy file used to describe the clarity of the LR images in each scene. Each HR and LR image has corresponding pixel quality indicator masks, namely SM and QM. They indicate which pixels are not covered by clouds, ice, snow, etc., and which pixels are able to be reliably used for reconstruction.

\subsection{Evaluation Metric}

The PSNR and SSIM are the primary evaluation metrics for general super-resolution, but not for the PROBA-V dataset. cPSNR~\cite{martens2019super} improved by PSNR and cSSIM based on SSIM are utilized as evaluation metrics for  MISR-RS.

\begin{table}
    \vspace{-0.3cm}
    \centering
    \caption{\textbf{Quantitative evaluation.} Comparison with the state-of-the-art methods on the two bands of the PROBA-V dataset, RED and NIR. The best results are marked in red, and the second best results are represented in blue}
    \begin{tabular}{ccccc}
    \toprule
      & \multicolumn{2}{c}{NIR}& \multicolumn{2}{c}{RED}  \\
    \cmidrule(lr){2-3}\cmidrule(lr){4-5}
        Method    & cPSNR & cSSIM & cPSNR & cSSIM\\
    \midrule
       Bicubic     & 45.44dB & 0.9770 & 47.33dB & 0.9840\\
    BTV ~\cite{farsiu2004fast}    & 45.93dB &  0.9794& 48.12dB &0.9840 \\
    IBP~\cite{irani1991improving}     & 45.96dB & 0.9796 & 48.21dB &09865 \\
    RCAN ~\cite{zhang2018image}    & 45.66dB & 0.9798 &48.22dB&0.9870 \\    
    VSR-DUF~\cite{jo2018deep}     & 47.20dB & 0.9850 & 49.59dB & 0.9902\\
    HighRes-net~\cite{deudon2019highres}    & 47.55dB & 0.9855 & 49.75dB & 0.9908\\
    3DWDSRNet~\cite{dorr2020satellite}     & 47.58dB & 0.9856 & 49.90dB & 0.9908\\
    DeepSUM ~\cite{molini2019deepsum}    & 47.84dB & 0.9858 & 50.00dB & 0.9908\\
    MISR-GRU ~\cite{arefin2020multi}    & 47.88dB & 0.9861 & 50.11dB & 0.9910\\
    DeepSUM++ ~\cite{Molini2020DeepsumND}    & 47.93dB & 0.9862 & 50.08dB & 0.9912\\
    RAMS ~\cite{salvetti2020multi}    & 48.23dB & 0.9869 & 50.17dB & 0.9913\\
    RAMS$+_{20}$ ~\cite{salvetti2020multi}    & 48.51dB & 0.9880 & 50.44dB & 0.9917\\
    TR-MISR ~\cite{an2022tr}    & 48.54dB & 0.9882 & {\color{blue}{50.67dB}} & 0.9921\\
    PIU  ~\cite{valsesia2021permutation}   & \color{blue}{48.72dB} &{\color{blue}{0.9883}} & 50.62dB & {\color{blue}{0.9921}}\\
    \textbf{ESC-MISR (ours)} & \textbf{\color{red}{49.42dB}}& \textbf{\color{red}{0.9896}}& \textbf{\color{red}{51.38dB}} & \textbf{\color{red}{0.9932}} \\
    \bottomrule
    \end{tabular}
    \captionsetup{font=small}

    \vspace{-0.3cm}
    \label{tab:compare}
\end{table}

\subsection{Experimental setting}
\indent
Our ESC-MISR takes slight longer to train because we need to calculate the message tokens of $N$ patches containing $n$ pixels and focus on reliable pixel features, where $N=\frac{H\times W}{n}$ and $n=1$. This is a significant computational burden for MIST, and we set the number of input LR images K to 24, determined by ESC-MISR (Fig.~\ref{fig:input_K}). We try to minimize the network size as much as possible to reduce parameters and accelerate the training speed. ESC-MISR does not require pre-training on other datasets, and the network weights are randomly initialized. In the feature encoding process, we only expand the number of channels from 2 to 32, which has been reduced by twice compared to the baseline. The embedding dimension of CMT is only 16, and 6 MISABs are applied in MIST.The batch size is 4 and training epochs = 400. We train each band on a NVIDIA GeForce RTX 3090 GPU.



\begin{figure*}
    \centering
    \includegraphics[width=0.95\linewidth]{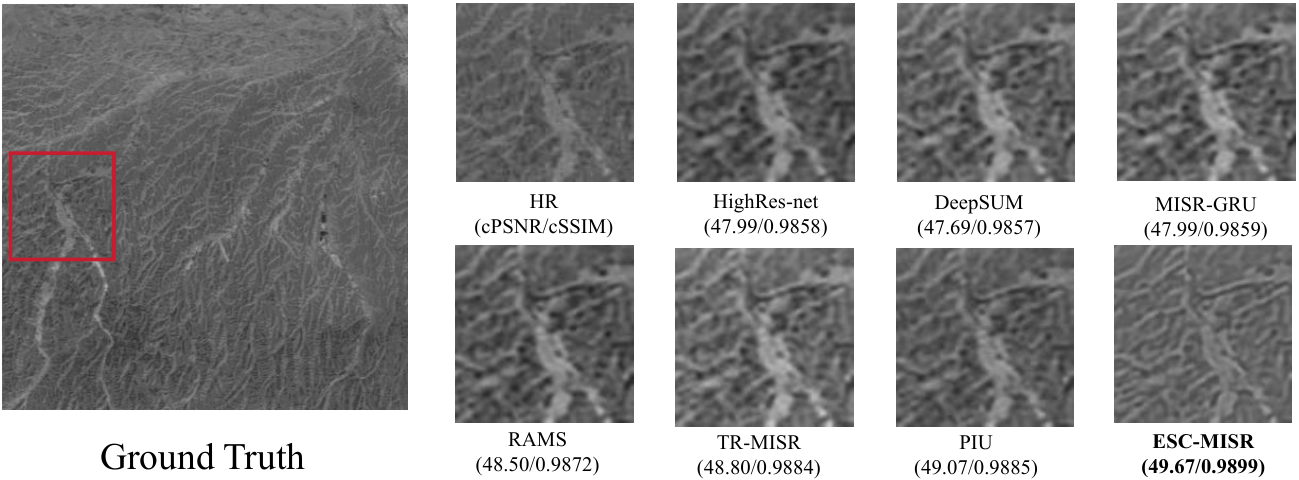}
    \captionsetup{font=small}
    \vspace{-0.3cm}
    \caption{\textbf{Qualitative evaluations}. Visual comparison between different MISR-RS methods on the imgset0614 scene of the NIR band. The comparative parts are highlighted with red boxes.}
    \vspace{-0.3cm}
        \label{fig:visual_NIR}
\end{figure*}

\begin{figure*}
    \centering
    \includegraphics[width=0.95\linewidth]{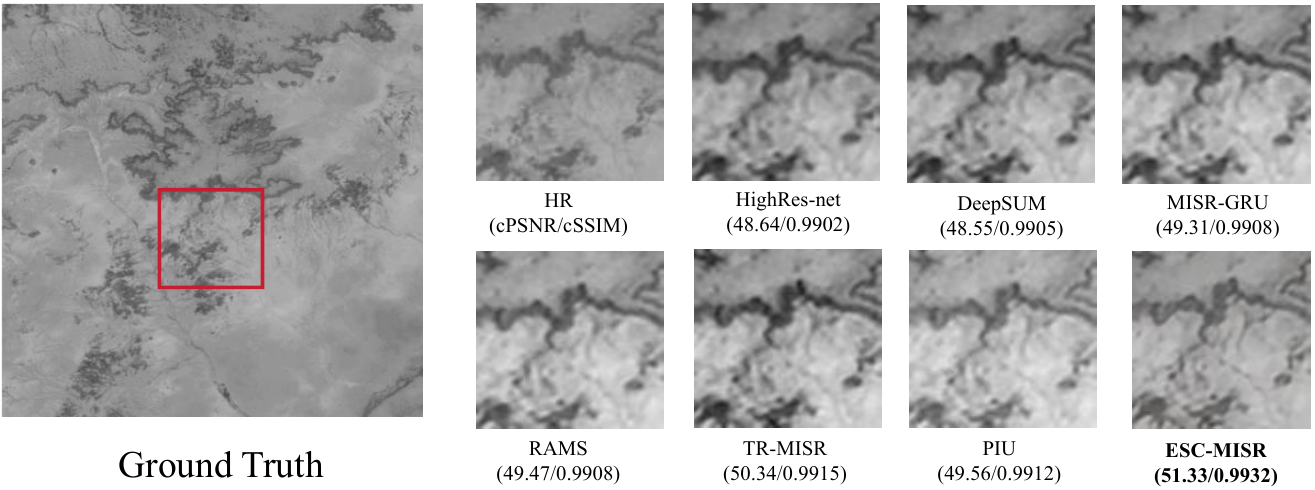}
    \captionsetup{font=small}
    \vspace{-0.3cm}
    \caption{\textbf{Qualitative evaluations}. Visual comparison between different MISR-RS methods on the imgset0024 scene of the RED band. The comparative parts are highlighted with red boxes.}
    \vspace{-0.6cm}
    \label{fig:visual_RED}
\end{figure*}

\subsection{Comparison with the State-of-the-Art Methods}
\noindent
\textbf{Quantitative Comparison.}
As presented in Tab.~\ref{tab:compare}, ESC-MISR is compared with Bicubic, BTV~\cite{farsiu2004fast}, IBP~\cite{irani1991improving}, RCAN~\cite{zhang2018image}, VSR-DUF~\cite{jo2018deep}, HighRes-net~\cite{deudon2019highres}, 3DWDSRNet~\cite{dorr2020satellite}, MISR-GRU~\cite{arefin2020multi}, DeepSUM++~\cite{Molini2020DeepsumND}, RAMS~\cite{salvetti2020multi}, RAMS$+_{20}$~\cite{salvetti2020multi}, TR-MISR~\cite{an2022tr} and PIU~\cite{valsesia2021permutation}, where Bicubic and RCAN are SISR methods, VSR-DUF is the Video Super-Resolution (VSR) method, and the rest is the MISR-RS method. Our method performs best in the NIR and RED bands of the PROBA-V dataset and is significantly superior to other methods. Specifically, ESC-MISR increases the cPSNR of PIU from 48.54dB to 49.42dB on the NIR band bringing an improvement of 0.88dB. It surpasses the current best result by 0.71dB on the RED band, which makes a breakthrough in MISR-RS tasks. These results demonstrate the effectiveness of our framework. {Our ESC-MISR can fully utilize spatial-temporal relations of LR images for SR.}

\noindent
\textbf{Visual Results.} 
We visualize the performance of ESC-MISR and other methods in reconstructing images on NIR and RED in the PROBA-V dataset, as displayed in Fig.~\ref{fig:visual_NIR} and Fig.~\ref{fig:visual_RED}. 
We randomly select three sets of images in the NIR and RED bands. Regardless of whether from a local or global perspective, our ESC-MISR reconstruction of the SR achieved great results. These results indicate that ESC-MISR can better focus on the clearer parts of global spatial features, fully utilize more reliable pixels, and generate more stable results for different frame sequences.

\subsection{Ablation Study}\label{ablation}

\begin{table}
\vspace{-0.6cm}
    \centering
        \caption{\textbf{Ablation study on each module of our ESC-MISR.} The best results are marked in red, and the second best result are represented in blue}
    \begin{tabular}{ccc|cc}
    \hline
         CMT&MIST&FFC&NIR cPSNR &RED cPSNR \\
     \hline   
        \usym{2717} & \usym{2717} & \usym{2717} &48.31dB&50.63dB\\
     \usym{2713} & \usym{2717} & \usym{2717} & 48.76dB &50.78dB  \\
     \usym{2717} & \usym{2713} & \usym{2717} & 48.60dB & 50.82dB \\
     \usym{2717} & \usym{2717} & \usym{2713} & 48.15dB  & 49.83dB \\
     \usym{2713} & \usym{2713} & \usym{2717} & {\color{blue}{49.12dB}} & {\color{blue}{51.07dB}} \\
     \usym{2713} & \usym{2717} & \usym{2713} & 48.86dB & 50.91dB \\
     \usym{2717} & \usym{2713}& \usym{2713} & 48.61dB & 50.78dB  \\
     \usym{2713} & \usym{2713} & \usym{2713} & \textbf{\color{red}{49.25dB}} & \textbf{\color{red}{51.21dB}} \\
    
      \hline       
    \end{tabular}
\vspace{-0.5cm}
    \label{tab:E2E}
\end{table}
\noindent
\textbf{Impact of Components of ESC-MISR.}
Tab.~\ref{tab:E2E} displays the quantitative performance of each component proposed by ESC-MISR. The effects of using CMT and MIST separately have been presented while using FFC alone can lead to poor results on NIR. This may be due to the poor image quality in NIR, as the combination of FFC and CMT or MIST can significantly improve the results. When MIST is used after CMT encoding, cPSNR shows a 0.54dB improvement in NIR compared to the case when MIST is not used. As presented in Tab.~\ref{tab:fusion module}, there is just a 0.29dB improvement compared to the other fusion blocks. 
 This demonstrates that a good encoder can assist the fusion module in reaching greater progress. The combination of CMT, MIST, and FFC in pairs can attain varying degrees of performance improvement. When three designs are combined, further improvement can be achieved, indicating the effectiveness of the overall design.


\begin{wraptable}{r}{7.5cm}
    \vspace{-1cm}
    \centering
        \caption{\textbf{Analysis of Fusion module}. Quantitative comparison on cPSNR between MISAB and other fusion blocks.}
    \begin{tabular}{ccc}
    \hline
    Fusion Blocks  & NIR cPSNR & RED cPSNR\\
     \hline
    GAP &47.24dB &49.58dB\\
        ConvGRU	&47.48dB	&49.74dB\\
        ConvLSTM&	47.45dB&	49.84dB\\
        Recursion&	47.03dB&	49.37dB\\
        3D Conv&	47.40dB	&49.71dB\\
        3D Conv+Attention&	47.79dB	&49.80dB\\
        self-attention &48.31dB&	50.63dB\\
\textbf{MISAB (ours)}	&\textbf{48.60dB}	&\textbf{50.82dB}\\
    \hline
    \end{tabular}
    \vspace{-0.5cm}
    \label{tab:fusion module}
\end{wraptable}
\noindent
\textbf{Effectiveness of MISAB.}
We conduct experiments to demonstrate the effectiveness of our MISAB by using different fusion blocks on the baseline. In order to reduce training time, we use TR-MISR~\cite{an2022tr} as the baseline. As shown in Tab.~\ref{tab:fusion module}, the performance of MISAB on NIR is 0.29dB-1.36dB better than other blocks in cPSNR, and 0.19dB-1.24dB higher on RED. Apparently, the fusion block based on attention is much better than other networks. Our MISAB uses image tokens with message tokens of pixels for cross-attention operations to enhance spatial correlations and avoid paying too much attention to temporal information. It has an advantage over normal attention blocks on MISR-RS tasks compared to the general fusion module that treats the input multi-image as temporal-related multiple frames.


\begin{wrapfigure}{r}{6.5cm}
    \centering
    \vspace{-0.3cm}
    \includegraphics[width=0.9\linewidth]{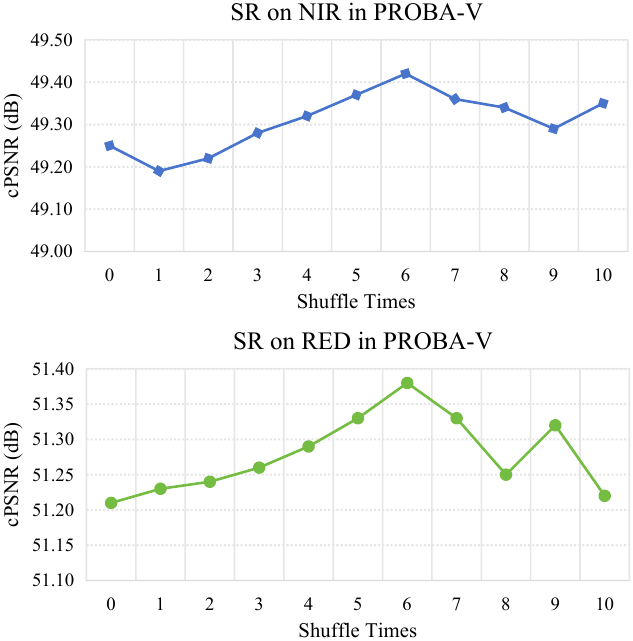}
    \captionsetup{font=small}
    \caption{\textbf{Impacts of shuffle times in random shuffle on cPSNR.} The figure illustrates that as the number of random shuffles increases, the performance of ESC-MISR will tend to stabilize}
    \vspace{-0.4cm}
    \label{fig:shuffle times}
\end{wrapfigure}
\noindent
\textbf{Influence of Shuffle strategy.}
We will discuss the effectiveness of the shuffle strategy in the training stage. ESC-MISR without random shuffle is treated as the baseline. We find that the effect of 1-time random shuffle is not as good as the baseline, and the improvements in NIR and RED can be observed from the third attempt of random shufﬂe. The reason for the analysis is that the randomness of the 1-time random shuffle is not enough, and ESC-MISR cannot adapt well to the interference caused by different orders on NIR. 
The discussion on the number of random shuffles is illustrated in Fig.~\ref{fig:shuffle times}. We perform random shuffles up to 10 times. On NIR, the performance of shuffles starts to improve after the third shuffle, while on RED, there is a slight improvement (0.02dB) after the first shuffle. At the sixth attempt, cPSNR reaches its highest improvement by 
0.17dB on NIR and RED, 
{proving that multiple random shuffles can indeed help the model weaken the interference of input orders and capture weak temporal correlations in LR images.}
    

\begin{wrapfigure}{r}{6.5cm}
    \centering
    \includegraphics[width=1\linewidth]{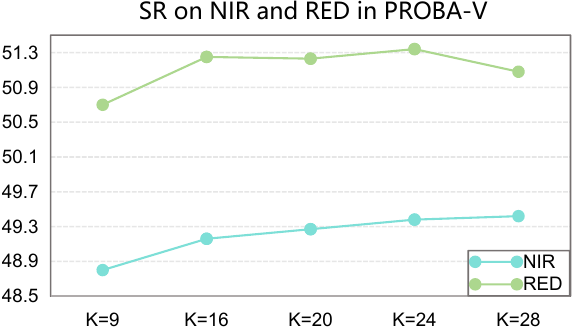}
    \captionsetup{font=small}
    \caption{{\textbf{Impact of the input images K}. cPSNR performance of the different numbers of K on SR for NIR and RED bands.}}
    \vspace{-0.4cm}
    \label{fig:input_K}
\end{wrapfigure}

\noindent
\textbf{Impact of the Number of Input Images.}
We discuss the impact of the number of input images K on our overall ESC-MISR. 
Unlike PIU~\cite{valsesia2021permutation}, RAMS~\cite{salvetti2020multi}, and DeepSUM~\cite{molini2019deepsum}, the performance of these model decreases with the increase of the number of input images. As shown in Fig.~\ref{fig:input_K}, ESC-MISR can fully utilize the effective information between multiple images when facing more images. On NIR, when K=28, cPSNR reaches 49.42dB, which is 0.05dB higher than the best result we achieved on K=24, but lower than when K=16 on RED. Analyzing the training process, we ﬁnd that over-ﬁtting occurred when K=28 when evaluation on RED. As K increases, the training time also increases. Therefore, we chose K=24 as the hyper-parameter, since ESC-MISR generalizes best on the validation set under that setting.

\section{Conclusion}
\label{sec:conclusion}

In this work, we have introduced the ESC-MISR framework that emphasizes the exploitation of spatial-temporal relations among low-resolution remote sensing images. Leveraging the CNNs-meet-transformers, the ESC-MISR framework encodes features more effectively and utilizes fast Fourier convolution in the decoding process to enhance the receptive field significantly. Furthermore, our fusion module MIST is specifically designed to capitalize on the complementary spatial information between images, thus facilitating superior high-resolution image reconstruction while minimizing temporal discrepancies. 
The random shuffle strategy for the low-resolution images in each scene has also proven critical. By reducing temporal dependencies, this strategy enables the model to capture weak temporal correlations effectively, which is often overlooked in traditional MISR methods. Our extensive experimental results affirm that ESC-MISR sets a new benchmark, surpassing existing state-of-the-art methods with notable improvements in cPSNR on standard datasets.

%
%
%
\bibliographystyle{splncs04}
%

\end{document}